\documentclass[letterpaper, 10 pt, conference]{ieeeconf}
\IEEEoverridecommandlockouts
\usepackage{graphicx}
\usepackage{float}
\usepackage{amsmath, amssymb, amsfonts}
\usepackage{cite}
\usepackage{caption}
\usepackage{subcaption}
\usepackage{hyperref}

\usepackage{graphics}
\usepackage{epsfig}
\usepackage{times}
\usepackage[dvipsnames]{xcolor}
\usepackage{booktabs}
\usepackage{multirow}
\usepackage{makecell}
\usepackage{pifont} 
\usepackage{colortbl}

\newcommand{\etal}{\textit{et al.}}

\title{\LARGE \bf
SEA: Semantic Map Prediction for Active Exploration of Uncertain Areas
}

\author{
    \begin{tabular}{c}
        \centerline{Hongyu Ding$^{1,*}$, Xinyue Liang$^{1,*}$, Yudong Fang$^{2}$, You Wu$^{1}$,} \\
        \centerline{Jieqi Shi$^{2,\dagger}$, Jing Huo$^{1,\dagger}$, Wenbin Li$^{2}$, Jing Wu$^{3}$, Yu-Kun Lai$^{3}$, Yang Gao$^{2}$}
    \end{tabular}
    \thanks{\textbf{$^{*}$Equal Contribution}, \textbf{$^{\dagger}$Corresponding Author}}
    \thanks{$^{1}$Hongyu Ding, Xinyue Liang, You Wu and Jing Huo are with the School of Computer Science, Nanjing University, China. Emails: {\texttt{\{hongyuding, MF21330051, you\}@smail.nju.edu.cn}, \texttt{huojing@nju.edu.cn}}}
    \thanks{$^{2}$Yudong Fang, Jieqi Shi, Wenbin Li and Yang Gao are with the School of Intelligence Science and Technology, Nanjing University, China. Emails: {\texttt{231880023@smail.nju.edu.cn}, \texttt{\{isjieqi, liwenbin, gaoy\}@nju.edu.cn}}}
    \thanks{$^{3}$Jing Wu and Yu-Kun Lai are with Cardiff University, United Kingdom. Emails: {\texttt{\{WuJ11, LaiY4\}@cardiff.ac.uk}}}
}

\begin{document}
\maketitle
\thispagestyle{empty}

\begin{abstract}
    In this paper, we propose SEA, a novel approach for active robot exploration through semantic map prediction and a reinforcement learning-based hierarchical exploration policy. Unlike existing learning-based methods that rely on one-step waypoint prediction, our approach enhances the agent's long-term environmental understanding to facilitate more efficient exploration. We propose an iterative prediction-exploration framework that explicitly predicts the missing areas of the map based on current observations. The difference between the actual accumulated map and the predicted global map is then used to guide exploration. Additionally, we design a novel reward mechanism that leverages reinforcement learning to update the long-term exploration strategies, enabling us to construct an accurate semantic map within limited steps. Experimental results demonstrate that our method significantly outperforms state-of-the-art exploration strategies, achieving superior coverage ares of the global map within the same time constraints. Project page: \url{https://robo-lavira.github.io/sea-active-exp/}
\end{abstract}


\section{INTRODUCTION}
    With the advancement of embodied agents, research on active exploration in unknown environments has grown rapidly. Exploration tasks in embodied intelligence differ in focus, including discovering and cataloging objects~\cite{chaplot2020semantic, ramakrishnan2021exploration}, locating specific targets~\cite{ye2021auxiliaryob, chaplot2020object, zhu2022navigating}, maximizing explored area within limited time~\cite{chaplot2020learning, bigazzi2022focus, juncheng2021learning}, and achieving accurate environment reconstruction.

    Traditional exploration algorithms rely on hand-crafted optimization methods to maximize coverage. Graph-based methods~\cite{dolgov2008practical} are precise but inflexible, often falling into local optima, while sampling-based methods~\cite{zhou2019robust,zhou2021fuel,Cao2021TAREAH} handle high-dimensional spaces but cannot guarantee optimality. These approaches generally face a trade-off between global awareness and exploration efficiency. To overcome this, reinforcement learning (RL) and deep learning (DL) have been introduced, leveraging neural networks’ implicit long-term memory to infer global structure~\cite{cao2024dare}. However, studies on spatial intelligence reveal that neural networks often fail to capture topology and lack robust 3D perception~\cite{yang2024thinking}, limiting their reliability in complex scenes.
    
	\begin{figure}[t]
		\begin{center}
			\includegraphics[width=0.98\linewidth]{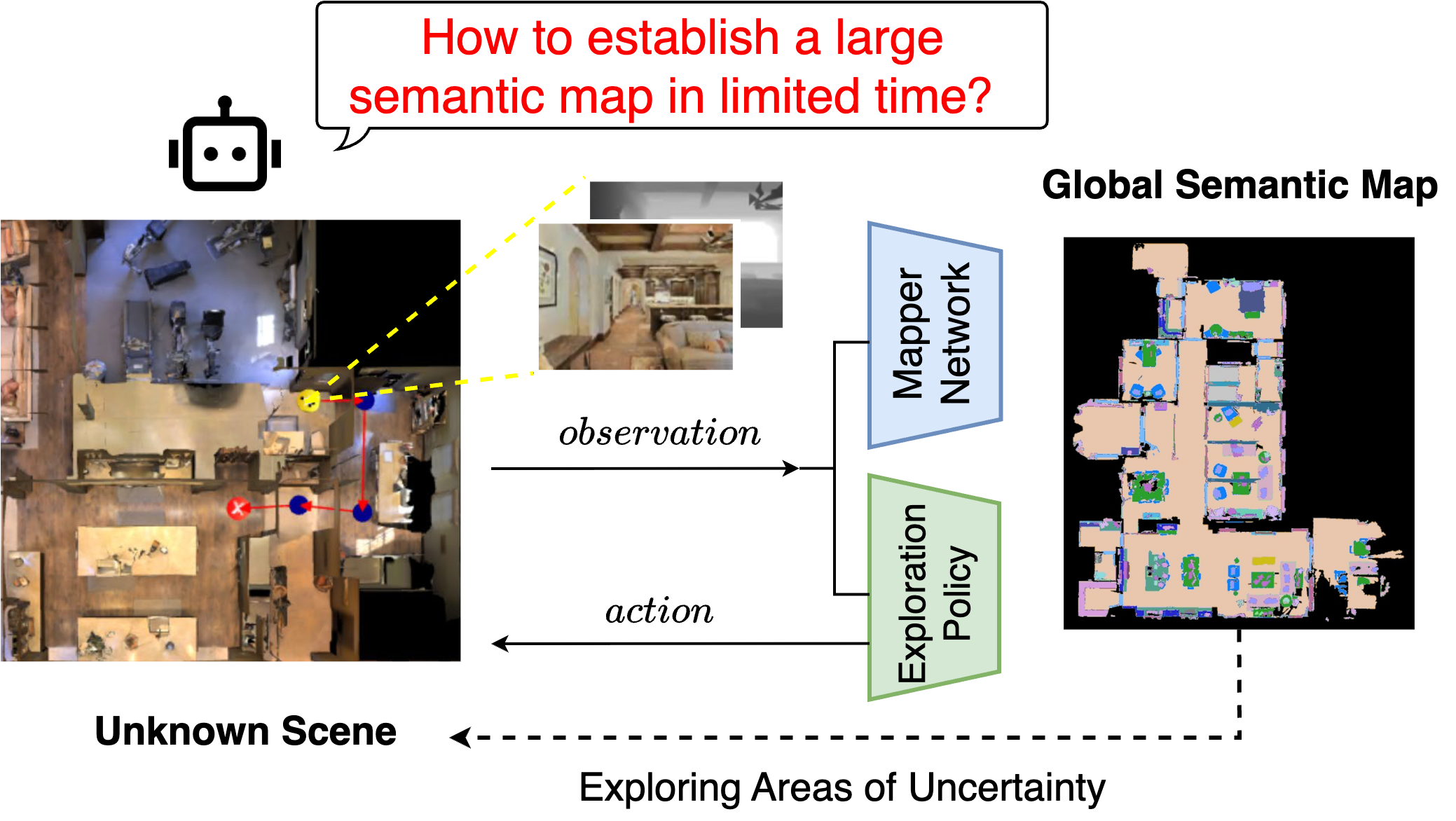}
		\end{center}
		\caption{Constructing a global semantic map within limited time steps, with the semantic map guiding active exploration. In our framework, we encourage the agent to take action to explore the uncertain areas on the semantic map to enhance exploration efficiency.}
		\label{fig:intro}
	\end{figure}
	
    To address these limitations, we integrate the adaptive planning capabilities of RL with explicit spatial reasoning to achieve more efficient and flexible exploration. Inspired by vision-and-language navigation (VLN), where semantic maps improve memory~\cite{Wang2023GridMMGM,Chen2022ThinkGA}, we argue that learning-based exploration should return to its core goal: actively and comprehensively covering unknown environments while building complete and accurate maps for decision-making.

The processes of explicit semantic map construction and unexplored area exploration operate in an iterative manner. The semantic map serves both as a structured representation of the environment and a form of spatial memory. It enables the agent to distinguish between explored and unexplored regions while continuously refining its action strategies for maximum efficiency. Simultaneously, the exploration process actively refines the map, ultimately ensuring the most rapid and comprehensive coverage of the environment. This interplay fosters an exploration strategy that balances exploration, which encourages enlarging the map, and correction, which encourages adding observations to refine uncertain areas. 

To achieve this, we train a prediction-based completion network that estimates complete maps from local observations, guided by prior knowledge of room layouts, object semantics, and shapes. A confidence-aware fusion module integrates observations into a global map, where discrepancies between predicted and accumulated maps highlight unexplored regions. By converting these unknown areas into a probabilistic map, we constrain the exploration policy to uncertain regions lacking prior knowledge, which are also areas we regard as more valuable and worth exploring. A tailored reward function further encourages accurate semantic coverage, improving waypoint selection and overall efficiency.

In summary, we propose \textbf{SEA}, a semantic-guided exploration framework that focuses on learning-based exploration and presents four key contributions: 1)We propose a prediction-based semantic map completion method to identify unexplored areas and guide long-term exploration strategies for faster map coverage. 2)We design a reward function that optimizes exploration waypoint selection by encouraging agents to prioritize uncertain regions. 3) Our approach efficiently explores complex environments while simultaneously generating an accurate global semantic map, enabling better navigation in unknown areas. 4)The proposed method outperforms state-of-the-art DRL-based exploration approaches on the Habitat dataset, demonstrating superior exploration efficiency and semantic map accuracy.

\begin{figure*}[ht]
    \begin{center}
        \includegraphics[width=0.85\linewidth]{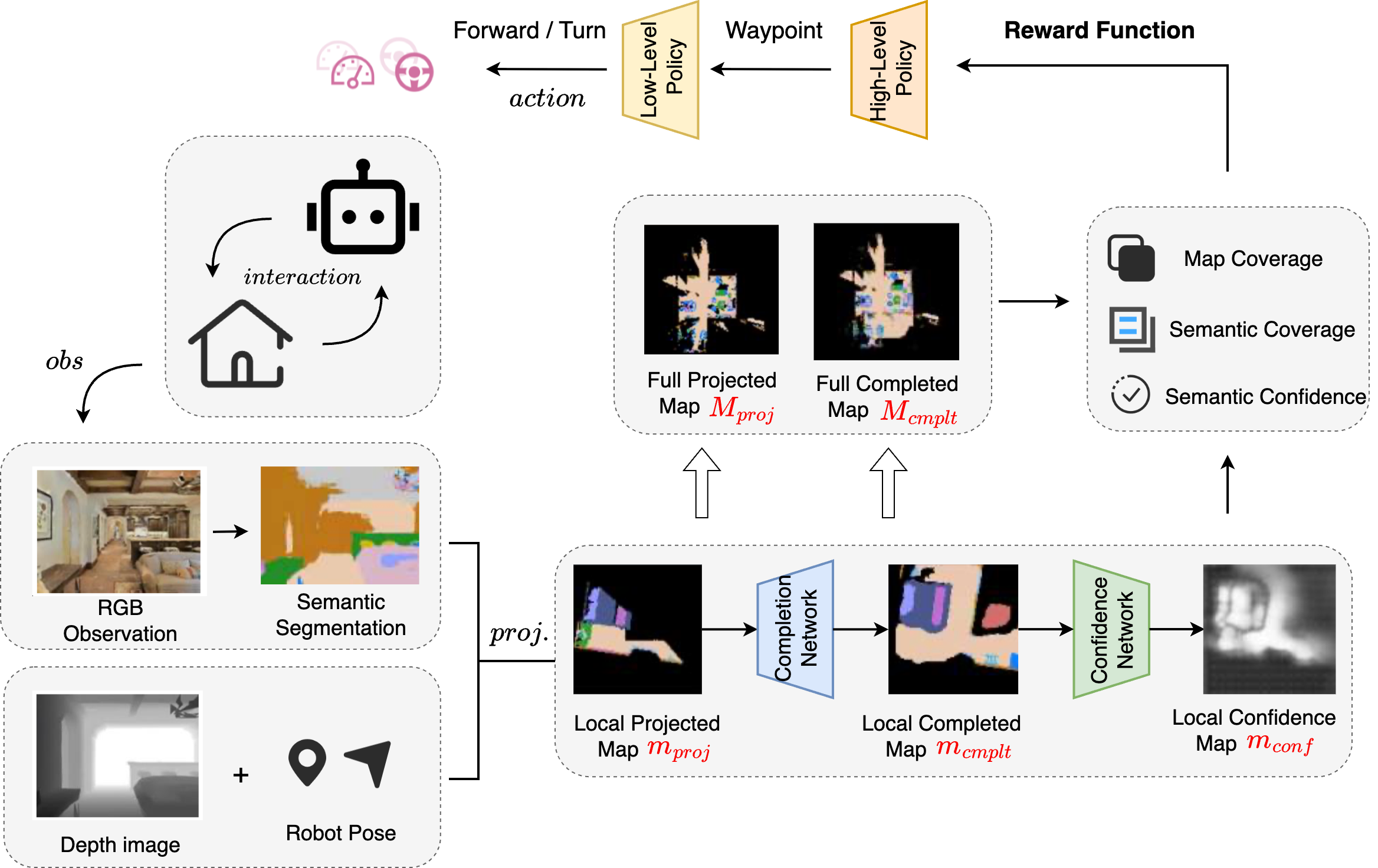}
    \end{center}
    \caption{\textbf{Overview of SEA.} After obtaining the egocentric observations of RGB-D image and the ground truth position from Habitat simulator as the input, we employ panoramic semantic segmentation and project the egocentric observations onto the 2D plane to build a local map. The projected local map is then processed by our Local Mapper and Full Mapper modules to generate local completed and confidence maps, which are subsequently accumulated into full projected and full completed maps. Besides, RL Navigator takes the local maps to select the long-term goal to explore, which is then passed to the short-term policy module to make the decision of the next action.}
    \label{fig:overview}
\end{figure*}

\section{Related Work}

\subsection{Semantic Scene Completion}
A concept closely related to our work is Semantic Scene Completion (SSC), which aims to infer the complete geometry and semantics of a scene from sparse inputs like depth maps or monocular RGB images. Early work primarily focused on object-level completion, reconstructing full geometry from limited 2D/3D observations~\cite{Yuan2018PCNPC,Tchapmi2019TopNetSP}. With the rise of deep learning, research has shifted toward completing full semantic scenes by jointly predicting semantics and geometry~\cite{Song2016SemanticSC,Li2023VoxFormerSV,jiang2023symphonies}. However, a key distinction from our approach is that most SSC methods reconstruct only the current visible scene in dense 3D space, without predicting unseen areas beyond the field of view.

Several studies combine completion with reconstruction and exploration~\cite{feng2023predrecon,Gao2024ActiveVP,Xu2023HeuristicbasedIP}. For instance, PredRecon~\cite{feng2023predrecon} applies object-level completion to guide UAV waypoint selection for reconstruction, yet it focuses on object geometry rather than scene-level exploration. The heuristic-based method in~\cite{Xu2023HeuristicbasedIP} identifies unknown frontiers and applies rule-based strategies to guide exploration. Although it integrates unknown-area prediction, it only distinguishes between explored and unexplored regions and lacks perceptual priors for active exploration.

\subsection{Occupancy Mapping and Semantic Mapping}
The objective of mapping is to construct a structured representation of an environment. Depending on whether semantic information is considered, this can be categorized into occupancy and semantic mapping. In occupancy mapping, Chaplot~\etal{} introduced an active navigation framework that projects egocentric views to generate obstacle maps~\cite{chaplot2020learning}. Other works construct obstacle maps~\cite{chen2019learning,ramakrishnan2020occupancy,ramakrishnan2021exploration,sharma2022occupancy} or build topological maps capturing spatial relations among rooms and objects~\cite{morad2021graph,wang2022lightweight,li2022remote,dang2022unbiased,chaplot2020neural,ravichandran2022hierarchical,zheng2021research}.

While some works~\cite{jayaraman2016look,jayaraman2018learning} attempt to enhance environmental observations by optimizing camera rotation for next-best-view selection, they do not guide the agent toward active exploration for improved semantic mapping. Similarly, studies on optimizing exploration perspectives for point cloud reconstruction~\cite{guedon2023macarons,cieslewski2017rapid,vasquez2014volumetric,vasquez2017view} adopt a next-best-view approach but are constrained by the high storage cost and computational resources required for 3D point clouds. As a result, many semantic mapping approaches struggle to integrate with autonomous exploration tasks.

\subsection{Active Exploration}
Active exploration aims to enlarge the explored area or discover more objects in an environment using limited steps. Our proposed task falls into this category but places a distinct emphasis on combining exploration with semantic map construction, requiring more explicit environmental modeling.

Most existing exploration approaches~\cite{chaplot2020learning,liu2021symmetry,juncheng2021learning,bigazzi2022focus,ramakrishnan2020occupancy,chaplot2020semantic,ramakrishnan2021exploration} focus on maximizing coverage or object diversity but lack explicit semantic reasoning. The works most related to ours are SSCNav\cite{liang2021sscnav} and L2M~\cite{georgakis2021learning}, which leverage predicted semantic local maps and uncertainty estimation. However, they are limited to object-goal navigation. SSCNav discards local maps at each step, preventing global map accumulation, and its confidence maps are used to correct errors rather than guide exploration in uncertain regions. L2M also exploits uncertainty, but only for specific object categories relevant to its navigation task, without extending to global semantic completion.

The recent work GLEAM~\cite{chen2025gleam} focuses on large-scale training and cross-dataset policy generalization. Its focus on scalability is complementary to ours; while GLEAM advances generalizable exploration, our method addresses the critical need for semantic uncertainty and completion to construct high-fidelity maps under a limited budget. To our knowledge, no prior work simultaneously explores unseen indoor environments and constructs predictive semantic maps to guide the process. We argue that semantic reasoning, fundamental to how humans explore, is a highly promising direction for enhancing both efficiency and scene understanding in embodied AI.

\section{Approach}

Our goal is to actively explore the environment while constructing an accurate semantic map within a limited number of steps. To achieve this, we propose a framework consisting of three key modules that facilitate both exploration and mapping. The overall approach is illustrated in Figure~\ref{fig:overview}. Specifically, we introduce three major components: (i) ASC-based Local Mapper that performs local map prediction and confidence estimation.
(ii) Two-stage Navigator that decouples the exploration task into long-term and short-term policies, allowing for more efficient and structured exploration. (iii) Confidence-aware Full Mapper that accumulates local maps while incorporating confidence-based adjustments.

\subsection{Task Definition}
In an unknown environment, the agent begins at a random position, actively exploring the scene while building a semantic map. At each timestep $t$, the agent acquires RGB-D observation $o_t$  and an accurate pose $p_t$ from the environment. The agent is supposed to learn an exploration policy $\pi(a_t| o_t,p_t)$ which aims to explore as rapid as possible while constructing an accurate semantic map by executing action $a_t \in\{MoveForward, RotateRight, RotateLeft\}$ within limited $k$ timesteps.

\subsection{ASC-based Local Mapper}
Building a complete map to represent the global environment requires fusing information from egocentric views. Existing DRL-based approaches \cite{li2023autonomous} directly accumulate maps by projecting only the areas visible to the agent, leading to inefficient mapping. We highlight that, with prior knowledge of indoor scenes, unseen areas can be predicted and completed, enhancing the semantic map with broader and more accurate coverage.

To address this, we propose the ASC-based Local Mapper, which projects local maps from egocentric observations and fills unseen areas with predicted semantics. To improve exploration efficiency, the agent is encouraged to explore regions with relatively inaccurate semantics. Additionally, the Local Mapper estimates semantic confidence and generates a confidence map to guide future exploration.

\begin{figure*}[ht]
\begin{center}
   \includegraphics[width=0.90
   \linewidth]{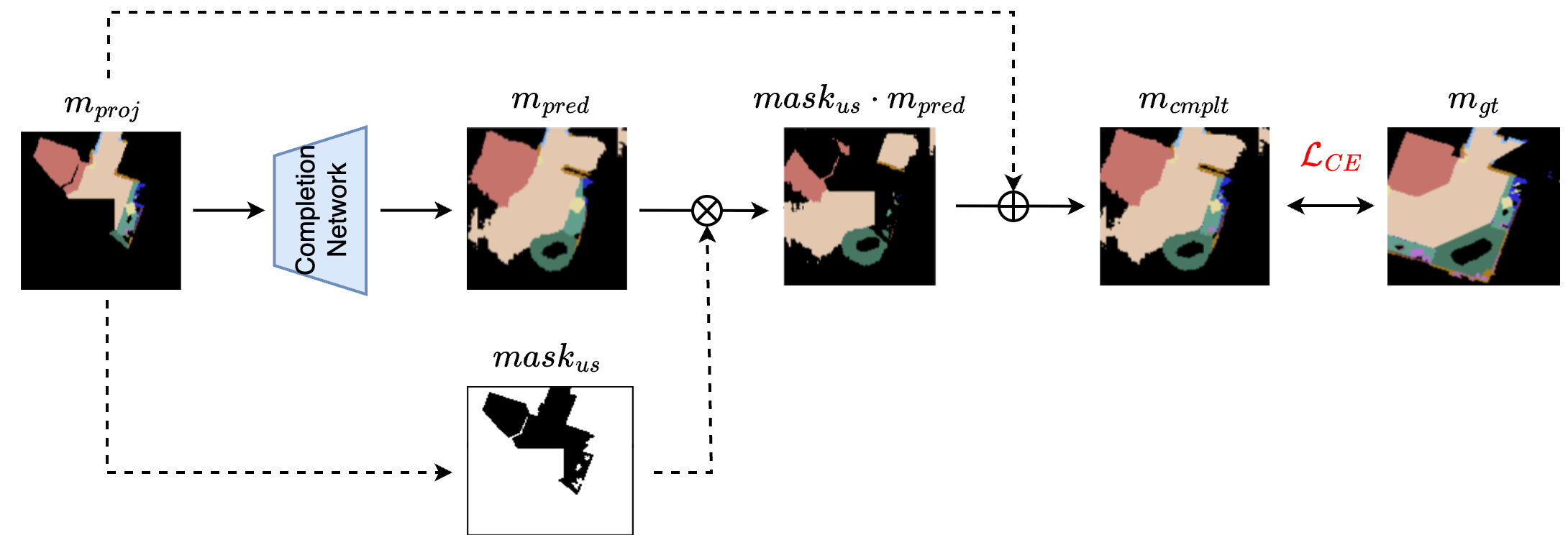}
\end{center}
   \caption{\textbf{The approach for building a local semantic completed map.} We pass the local projected map $m_{proj}$ through the semantic completed network to get the semantic prediction map $m_{pred}$, and extract the unseen mask $mask_ {us}$ from the unseen part. As shown in Equation~\ref{equ:mask}, we fill the unseen part of $m_{proj}$ with that in $m_{pred}$ to form $m_{cmplt}$. }
\label{fig:localcmplt}
\end{figure*}
 
 \subsubsection{Panoramic Segmented Local Map}

In this module, we first perform panoramic segmentation of RGB observation according to the 40 semantic categories of the MP3D dataset. Our panoramic semantic segmentation employs off-the-shelf ACNet~\cite{hu2019acnet} and leverages the pretrained panoramic segmentation model from SSCNav~\cite{liang2021sscnav} which is trained with 209,200 RGB-D images of 40 object categories from MP3D training houses. As a comparison, SemExp~\cite{chaplot2020object} utilizes only 16 semantic categories relevant to the Habitat Object Goal Navigation task. In contrast, our local maps include all 40 semantic categories plus an additional void category. Moreover, our method provides a panoramic semantic view, excluding unseen areas such as floors and walls.

We combine depth observation and segmented RGB observation to compute each point in the egocentric 3D point cloud. To project the 3D map into 2D space,  we accumulate the semantic point cloud based on height and obtain the 2D semantic local projected map $m_{proj}$ of size $ M\times M$. In addition, we determine the world coordinates of this 2D local map using the ground-truth rotation and translation of the agents. 

We emphasize that, although we do not explore real-world data expansion in this paper, we have aimed to replicate real-world noise conditions in this section to facilitate future research. To align with the semantic noise encountered in real-world applications, we employ a segmentation model for semantic segmentation instead of using the ground-truth annotations. Regarding pose estimation, while we obtain ground-truth poses directly from the simulator, this process can be replicated in the real world using a SLAM front end. Considering that common front-end methods like VINS can achieve an accuracy of 7 cm ATE in a 147 m trajectory\cite{Campos2020ORBSLAM3AA}, we believe this assumption is reasonable.

\subsubsection{Local Map Completion}
By leveraging prior knowledge of indoor scenes and objects, unseen areas can be predicted and completed based on the projected map, expanding accurate semantic coverage within a limited number of steps and improving exploration efficiency. In our framework, we generate a completed semantic local map, covering the entire local map and filling unknown regions with predicted semantics. Unlike SSCNav’s completion method, our approach maintains consistency between observed and predicted areas during training.

As illustrated in Figure~\ref{fig:localcmplt}, we calculate the unseen area in $m_{proj}$ to obtain a mask $mask_{us}  \in\{0,1\}^{M \times M}$. We exploit ResNet with a fully convolutional neural network with 1 maxpooling layer, 4 down-sample residual blocks, and 5 up-sample residual blocks, following the same architecture in SSCNav \cite{savva2019habitat}. Differently, we use only one observed projected map $m_{proj}$ to get the predicted map $m_ {pred}$ of the same size, and fill the unseen part of $m_{proj}$ with that in $m_{pred}$ to form $m_{cmplt}$ for consistency. The loss is calculated as the Cross Entropy Loss between $m_{cmplt}$ and $m_{gt}$.
\begin{equation}
	m_ {cmplt}= (1-mask_ {us})\odot m_{proj}+mask_ {us}\odot m_ {pred}
 \label{equ:mask}
\end{equation}
\subsubsection{Local Map Confidence Estimation}
If the predicted area of $m_{cmplt}$ lacks accuracy, semantic noise may introduce errors into the projected map, negatively affecting subsequent exploration. Since our goal is to maximize Accurate Semantic Coverage (ASC) within a limited number of steps, repeatedly exploring areas where the completed semantic map is already accurate or revisiting previously explored locations would be inefficient, limiting the expansion of full scene coverage.

To address this, we introduce a novel method that prioritizes areas where the completed semantic map is less accurate. We select such maps as high-potential regions for the next long-term exploration target. The agent is then guided by this target to explore these areas efficiently. In summary, since inaccurate regions require further exploration, we predict a semantic confidence map $m_{conf}  \in[0,1]^{M \times M}$ for $m_ {cmplt}$ to assess confidence levels and guide exploration.

We compare $m_{cmplt}$ and the ground truth local map $m_{cmplt}$ to generate a semantic accuracy map $m_{acc}\in\{0,1\}^{M \times M}$, which serves as supervision for the predicted confidence map. Using the same ResNet-based architecture as the completion network, we take $m_{pred}$ as the input during training and output $m_{conf}$. We only apply the Cross Entropy Loss of the unseen part between prediction map for optimization. During navigation, repeated exploration can be avoided by using $m_ {conf}$ as a guidance. With this approach, the RL Navigator based on $m_{conf}$ can efficiently construct a more accurate semantic map within a limited number of steps.

\subsection{Two-Stage Navigator}
We divide the exploration task into two distinct stages. First, we select a long-term goal to guide the agent’s navigation over the next $\mu$ steps. Then, we apply short-term policies for path planning and action selection, ensuring efficient movement toward the goal.

\subsubsection{Confidence-aware Long-term Policy}
Existing approaches~\cite{chaplot2020learning, chaplot2020semantic} select long-term goals across the entire map. However, our experiments show that these methods often select goals near map corners, limiting their ability to utilize environmental information effectively. This inefficiency negatively impacts the overall exploration process.

To address these limitations, we propose leveraging the Soft Actor-Critic (SAC) algorithm~\cite{haarnoja2018soft}, an off-policy, model-free reinforcement learning method that enables more effective long-term goal selection. Within a single long-term goal period, the agent's reachable distance remains limited to the ${M \times M}$ local map. This allows us to select a new long-term goal based on the current $m_{proj}$, $m_{cmplt}$ and $m_{conf}$. The interval for long-term goal planning is set to $\mu=25$. The outputs of the Local Mapper, including $m_{proj}$, $m_{cmplt}$ and $m_{conf}$, are fed into the SAC module for exploration. Additionally, the embeddings of the agent's current position and orientation are included as inputs.

Meanwhile, the Confidence-Aware Full Mapper (Subsection III.D) updates the full projected map. The differences between the coverage $\Delta_{cover}$ of the full projected maps is computed to determine the coverage reward $r_{cover}$. Since the predicted map area can theoretically expand indefinitely, maximizing its coverage alone is not meaningful. Therefore, we use the variance in Accurate Semantic Coverage (ASC) between consecutive steps of the completed map to compute the completion reward $r_{asc}$. Besides, we calculate the confidence of the long-term goal $r_{conf}$ on $m_{conf}$.

At time $t$, we define the following reward $r_t$ in reinforcement learning training, where $\eta_{c}$, $\eta_{asc}$ and $\eta_{conf}$ are hyper parameters that balance different loss terms,
\begin{equation}
\begin{aligned}
	r_{t}&=r_{cover}+r_{asc}+r_{conf}\\
 & =\eta_{c} \Delta_{c} +   \eta_{asc} \Delta_{asc}+\eta_{conf}  conf_{t}
 \label{reward}
\end{aligned}
\end{equation}
where $r_{cover}$ is the change in projected map coverage ($\Delta_{c}$), $r_{asc}$ is the change in Accurate Semantic Coverage of the completed map ($\Delta_{asc}$), and $r_{conf}$ is the confidence value of the chosen goal. The hyperparameters $\eta_{c}$, $\eta_{asc}$, and $\eta_{conf}$ balance these terms.

\subsubsection{Short-term Policy}
For the short-term policy, a deterministic path planning method, Fast Marching Method (FMM)~\cite{sethian1996fast}, is used. Based on $m_{proj}$, FMM plans a path toward the selected long-term goal and determines the agent’s next action. It maintains an obstacle local map, allowing the agent to navigate efficiently while avoiding obstacles during exploration.

\subsection{Confidence-aware Full Mapper}
Based on the output of the Local Mapper(Section III.B), we construct the full projected map $M_ {proj}$, full completed map $M_ {cmplt}$ and full confidence map $M_ {conf}$ to obtain the semantic map of an indoor scene. Firstly, we establish $M_ {conf}$ by accumulating local confidence map $m_{conf}$ according to the position. If a region overlaps with a previously explored area, we retain the higher confidence value. We define this process as confidence-aware accumulation. Simultaneously, we generate a mask for the retained parts of the current $m_{conf}$, denoted as $mask_{conf}$.

We then apply $mask_{conf}$ to $m_{proj}$ and $m_{cmplt}$ before stitching them into $M_{proj}$ and $M_{cmplt}$ according to the world coordinates of the local maps.
Finally, we evaluate the metrics using $M_{proj}$ and $M_{cmplt}$.

\section{Experiment Results}
\subsection{Experiment Setup}

\textbf{Dataset and Simulator.} We use Habitat~\cite{savva2019habitat} as the simulation platform and MP3D~\cite{chang2017matterport3d} as the dataset. MP3D is an RGB-D indoor dataset with 90 large-scale scenes: 61 for training, 11 for validation, and 18 for testing. Our method is trained with 1M observation frames from the training split and evaluated with 50k frames from the validation split. For semantic completion and confidence modeling, we adopt the local map completion dataset from SSCNav, which contains over 52.3k pairs of ground-truth and projected maps. Ground-truth full maps $M_{gt}$ for evaluation are generated following SMNet~\cite{cartillier2021semantic} using MP3D semantics.

\textbf{Metrics.} To assess the effectiveness of semantic mapping, we define the following metrics. \textbf{CovP} is the area covered by  $M_{proj}$ in $m^2$ and \textbf{CovC} is the coverage of $M_{cmplt}$ in $m^2$. \textbf{ASCP} is accurate semantic coverage which is calculated as the interaction in $m^2$ between $M_{proj}$ and $M_{gt}$ while \textbf{ASCC} uses $M_{cmplt}$. 

Following~\cite{bigazzi2022focus}, the maximum exploration step is set to 500. The local map resolution is $128\times128$, and the full map is $1280\times1280$ (corresponding to $48\times48,m^2$ in the physical world). Although insufficient for full exploration, this effectively highlights differences across strategies for semantic map construction. Long-term goals are updated every 10 steps. Empirically, ASC is about half of total coverage. We estimate $\Delta_{asc}$ by doubling it and compute $\Delta_{conf}$ by scaling $\Delta_{asc}$ by 1000. The final reward (Eq.~\ref{reward}) is calculated with hyperparameters $\eta_{c}=0.6$, $\eta_{asc}=0.4$, and $\eta_{conf}=0.4$.

\begin{table}[t]
\centering
\caption{Results after 500 steps on MP3D datasets.}
\label{res}
\setlength{\tabcolsep}{10pt}
\footnotesize
\begin{tabular}{l|cccc}
\toprule
\textbf{Method} & \makecell{\textbf{CovP} \\ ($m^2$) $\uparrow$} & \makecell{\textbf{ASCP} \\ ($m^2$) $\uparrow$} & \makecell{\textbf{CovC} \\ ($m^2$) $\uparrow$} & \makecell{\textbf{ASCC} \\ ($m^2$) $\uparrow$} \\
\midrule
SemExp* & 96.61 & 36.54 & 101.64 & 40.92 \\
Impact & 104.53 & 42.19 & 111.53 & 45.14 \\
EE & 102.79 & 40.95 & 109.47 & 46.48 \\
SSCNav & 14.61 & - & 14.61 & - \\
ARiADNE* & 104.25 & - & 104.25 & - \\
\midrule
\textbf{SEA (Ours)} & \textbf{111.74} & \textbf{49.53} & \textbf{117.14} & \textbf{53.99} \\
\bottomrule
\end{tabular}
\end{table}

\begin{table}[t]
\centering
\caption{Ablation results on MP3D Val datasets.}
\label{tab:val}
\setlength{\tabcolsep}{10pt}
\footnotesize
\begin{tabular}{l|cccc}
\toprule
\textbf{Method} & \makecell{\textbf{CovP} \\ ($m^2$) $\uparrow$} & \makecell{\textbf{ASCP} \\ ($m^2$) $\uparrow$} & \makecell{\textbf{CovC} \\ ($m^2$) $\uparrow$} & \makecell{\textbf{ASCC} \\ ($m^2$) $\uparrow$} \\
\midrule
w/o $r_{cover}$ & 104.20 & 43.10 & 106.36 & 44.15 \\
w/o $r_{asc}$ & 111.03 & 48.76 & 115.08 & 51.66 \\
w/o $r_{conf}$ & 110.59 & 48.07 & 114.74 & 50.49 \\
w/o $m_{cmplt}$ & 109.11 & 47.87 & 112.29 & 49.10 \\
w/o $m_{conf}$ & 106.63 & 45.80 & 108.18 & 46.88 \\
\midrule
\textbf{SEA} & \textbf{111.74} & \textbf{49.53} & \textbf{117.14} & \textbf{53.99} \\
\bottomrule
\end{tabular}
\end{table}

\begin{figure*}[ht]
\begin{center}
   \includegraphics[width=0.98\linewidth]{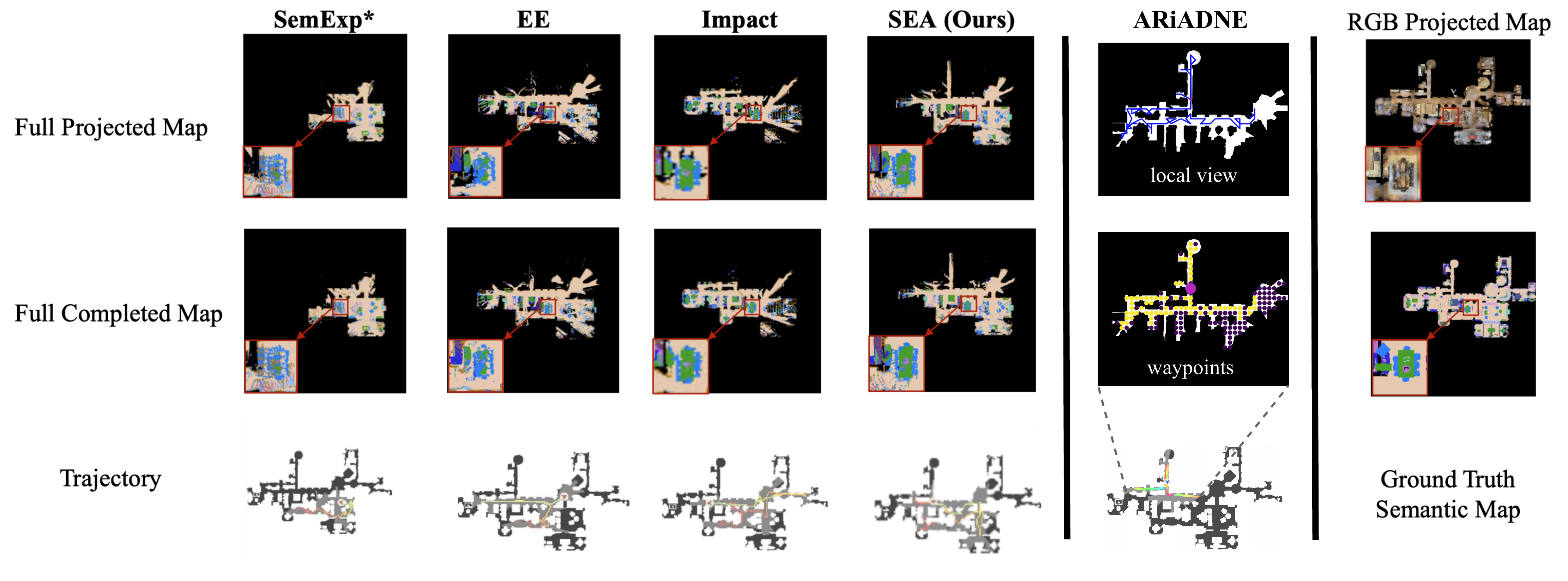}
\end{center}
   \caption{The qualitative evaluation of different methods during evaluation, including the full projected map and the full completed map. To provide a clearer view, we enlarge $m_ {proj}$ and $m_ {cmplt}$ built at a specific time point in the bottom left corner. Since ARiADNE has no semantic maps, we use different views according to its original paper for illustration.}
\label{fig:reuslt}
\end{figure*}

\subsection{Baselines}
In our simulation experiments, the agent’s pose is obtained directly from Habitat’s ground-truth position sensor, avoiding pose estimation and simplifying mapping. This allows us to focus on evaluating exploration speed, coverage integrity, and semantic certainty. Accordingly, we select baselines that emphasize map coverage.

\begin{itemize}
\item SemExp* \cite{chaplot2020object}: Originally designed for object-goal navigation, SemExp shares its architecture with ANS\cite{chaplot2020learning}, which targets occupied coverage. We modify its reward to prioritize coverage and retrain it on the same dataset, effectively converting it into an exploration method.

\item Impact \cite{bigazzi2022focus}: Trained with purely intrinsic rewards, Impact explores to maximize occupied coverage. We evaluate it using the provided trained model under the same setting reported in the original paper.

\item EE \cite{ramakrishnan2021exploration}: This work analyzes how reward design affects coverage and object detection. We adopt the released model with the best coverage performance for evaluation.

\item ARiADNE \cite{Cao2023ARiADNEAR} is a state-of-the-art RL-based exploration method that outperforms classical approaches like TARE \cite{Cao2021TAREAH} on grid maps. For a fair comparison, we adapted it to match our method's mapping scope and integrated our short-term policy, FMM, as its fine-grained exploration strategy in the continuous Habitat environment.

\end{itemize}

\subsection{Results on MP3D Datasets}
As shown in Table~\ref{res}, our method outperforms existing approaches in both global map coverage and semantic accuracy, for both projected and completed maps. After 500 steps, it achieves 111.74 $m^2$ of occupied coverage and 49.53 $m^2$ of accurate semantic coverage on the projected map.

ARiADNE, originally trained in simple, flat environments with regular layouts, performs poorly on MP3D, which contains diverse and irregular structures, often with multiple floors. For fairness, we report ARiADNE* on single-level scenes, where its performance is comparable to other baselines. These results highlight the advantage of our method in the selection of routes and the prediction of long-term semantics. By avoiding redundant exploration in already confident areas, it reallocates effort to uncertain regions, leading to more efficient exploration.

Notably, the best coverage reported in Impact \cite{bigazzi2022focus} when evaluated on the MP3D Val split is 144.64 $m^2$. However, this difference arises because our experimental setup limits the depth range of map projection to 5 m, whereas the mapper in Impact's code uses 10 m. Consequently, the coverage of our semantic map mapper appears lower than in Impact's original paper. Nevertheless, since all experiments are conducted under a consistent setting in our study, the results remain valid and reliable.

\begin{figure*}[ht]
\begin{center}
   \includegraphics[width=0.98\linewidth]{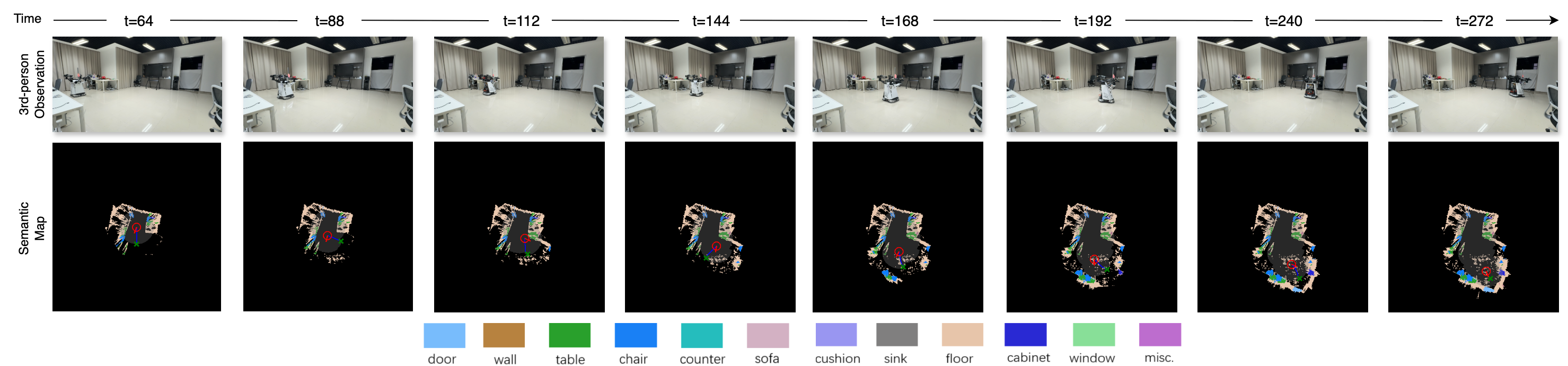}
    \end{center}
   \caption{The qualitative evaluation of SEA in a real-world setting. The top row shows third-person views of the robot's navigation at different timesteps. The bottom row displays the corresponding semantic mapping results. In the semantic maps, the red circle indicates the robot's pose and orientation, the green cross marks the next waypoint predicted by our model, and the blue line represents the planned path to that waypoint. The legend at the bottom explains the color-coding for different object categories in the semantic map.}
\label{fig:realw}
\end{figure*}

Figure~\ref{fig:reuslt} further illustrates agent trajectories in an unseen MP3D scene. Our method yields the widest coverage and highest semantic accuracy. The close-up around the dining table shows how multi-view observations produce a more precise semantic map, surpassing SemExp*, EE, and Impact in detail and correctness.

Furthermore, to demonstrate the effectiveness of each module in our proposed method, we conducted a comprehensive ablation study. Specifically, we systematically removed each component from the global input of the RL navigator as well as individual reward mechanisms. As quantitatively presented in Table \ref{tab:val}, our experimental results demonstrate that each module contributes significantly to the semantic map construction process. The most substantial performance degradation was observed when $m_{cmplt}$ and $m_{conf}$ were excluded from the global input, indicating that the superior performance primarily stems from the map-based inputs to the reinforcement learning model for long-term goal selection, rather than relying solely on feedback rewards. Moreover, our analysis reveals that $m_ {conf}$ and $r_ {conf}$, which specifically identify and quantify the semantically uncertain regions in the currently explored map, thereby enabling more effective optimization of long-term goal selection. The differential impact between these components suggests that the explicit representation of uncertainty plays a crucial role in guiding the exploration process.

\subsection{Real-world Deployment}
To further validate the practicality of our approach, we conducted a real-world experiment in an indoor office environment using the Agilex Cobot Magic mobile platform. The robot was equipped with an Intel RealSense D435i camera at the front, and we utilized the robot's native wheel odometry for localization. The system demonstrated high efficiency, with an average inference time of 0.08s for waypoint prediction and an average time of 0.75s per frame for semantic map construction. Within a fixed number of steps, our method enabled the robot to actively explore the environment, producing a semantic map that captured both the structural layout and object categories with high fidelity. This demonstrates its robustness and effectiveness beyond simulation. Representative results of the constructed semantic map in the office environment are shown in Figure~\ref{fig:realw}.

\section{Conclusion}
In this work, we introduced SEA, a visual exploration framework that prioritizes low-confidence regions on the semantic map to achieve rapid semantic coverage. By leveraging reinforcement learning for long-term goal selection, SEA guides the agent toward uncertain areas, enabling efficient observation of semantic objects, refinement of the map, and effective exploration of unseen environments within limited steps. Experimental results show that SEA consistently outperforms existing DL- and DRL-based exploration methods in both semantic map construction and indoor exploration.

\textbf{Limitations and Future Work.} Our study focused on predicting completed maps from local projections within a $4.8\times4.8m^2$ region around the agent. Extending completion and confidence estimation to the global map remains challenging due to model size, runtime constraints, and the mismatch between long-term goal selection and feasible movement ranges. Additionally, noise artifacts caused by distant objects beyond the simulator’s depth range highlight the need for more effective denoising methods that preserve semantic accuracy. We believe addressing these challenges will further advance semantic exploration and support broader applications of embodied AI in real-world environments.

\bibliographystyle{IEEEtran.bst}
\bibliography{refs}
\end{document}